\documentclass[lettersize,journal]{IEEEtran}
\usepackage{amsmath,amsfonts}
\usepackage{algorithmic}
\usepackage{algorithm}
\usepackage{array}
\usepackage[caption=false,font=normalsize,labelfont=sf,textfont=sf]{subfig}
\usepackage{textcomp}
\usepackage{stfloats}
\usepackage{url}
\usepackage{verbatim}
\usepackage{graphicx}
\usepackage{cite}
\usepackage{placeins}
\usepackage{float}

\usepackage[colorlinks=true,linkcolor=blue,citecolor=blue,urlcolor=blue]{hyperref} 
\usepackage{mathtools}
 \usepackage{pifont}               
\usepackage[dvipsnames,table]{xcolor}        
\usepackage{afterpage}
\colorlet{CMarkColor}{Green}
\colorlet{XMarkColor}{BrickRed}
\DeclareRobustCommand{\cmark}{{\color{CMarkColor}\ding{51}}} 
\DeclareRobustCommand{\xmark}{{\color{XMarkColor}\ding{55}}} 

\definecolor{rowhl}{HTML}{FFF3CD}   
\usepackage{tabularx,booktabs,array}
\newcolumntype{Y}{>{\centering\arraybackslash}X} 

\newcommand*\systemname{{\bf UniRED}}

\begin{document}

\title{UniRED: Unified RGB-D Video Frame Interpolation\\with Event Guidance}

\author{Yinuo Zhang, Guangshun Wei, Yuanfeng Zhou, Yiran Shen,~\IEEEmembership{Senior Member,~IEEE}
\thanks{Yinuo Zhang, Guangshun Wei, Yuanfeng Zhou, Yiran Shen are with the School of Software, Shandong University, Jinan 250100, China (e-mail: vermonth1216@gmail.com; guangshunwei@gmail.com; yfzhou@sdu.edu.cn; yiran.shen@sdu.edu.cn)}
\thanks{Corresponding author: Guangshun Wei and Yiran Shen}}

\markboth{Journal of \LaTeX\ Class Files,~Vol.~14, No.~8, August~2021}%
{Shell \MakeLowercase{\textit{et al.}}: A Sample Article Using IEEEtran.cls for IEEE Journals}


\maketitle
%
  
\begin{abstract}
High frame-rate RGB-D videos are crucial for a variety of downstream tasks, including motion analysis, dynamic scene understanding, and 3D reconstruction. However, due to hardware and sensing constraints, practical RGB-D cameras are typically limited to low frame rates, making it difficult to capture rapid scene dynamics. Existing video interpolation methods have achieved strong performance on RGB data, but they are not readily applicable to RGB-D scenarios, where they often yield blurry boundaries, visible artifacts, and degraded geometric consistency.
Furthermore, motion estimation from only two boundary frames is inherently under-constrained in complex dynamic scenes. Event cameras, by contrast, provide asynchronous measurements with ultra-high temporal resolution, offering dense motion cues.
In this paper, we propose a unified multimodal framework for RGB-D video interpolation that jointly exploits RGB appearance, depth geometry, and event-based temporal cues. 
Specifically, it first extracts and fuses RGB, depth and event cues, then estimates bidirectional flow with motion basis refinement for RGB and Z-axial refinement for depth, and finally synthesizes the target RGB-D frame via bidirectional warping and soft blending. 
In addition, we construct a new RGB-D-Event dataset to alleviate the scarcity of tri-modal training data.
Extensive experiments on a public benchmark and the proposed dataset demonstrate that our method achieves superior photometric fidelity for RGB interpolation and stronger geometric accuracy for depth interpolation than existing approaches.
\end{abstract}


\section{Introduction}
\IEEEPARstart{R}{GB-D} sensing augments conventional color imagery with per-pixel depth, providing temporally synchronized appearance and metric geometry \cite{handa2012real}.
This complementary representation has been widely adopted for accurate motion estimation, geometry-aware dynamic scene understanding, and photorealistic 3D scene reconstruction \cite{newcombe2015dynamicfusion,fu2018texture,10974473}.
However, practical RGB sensors are constrained by 
low frame rate (typically 30-50 fps) that restricts their ability to capture fine-grained motion dynamics, particularly in highly dynamic scenes with rapid object or camera motion, while depth sensors, such as mechanical LiDARs, operate at even lower frequencies (often around 10 Hz) \cite{geiger2013vision, caesar2020nuscenes}, causing severe temporal discontinuities.
Consequently, recovering high-quality intermediate RGB-D frames holds significant potential for improving perception continuity and geometric fidelity in dynamic environments.

Recently, video frame interpolation for RGB has been extensively studied in the computer vision community~\cite{kye2025acevfi}.
Most existing approaches reconstruct intermediate frames from two temporal boundaries through kernel-based resampling or optical flow–guided warping \cite{niklaus2017video1,niklaus2017video2,lee2020adacof,jiang2018super,huang2022real,niklaus2020softmax}.
In contrast, depth frame interpolation remains relatively underexplored.
Early methods largely rely on handcrafted priors and analytical optimization \cite{dolson2010upsampling,zhang2014texture,yuan2018temporal}, which often fail under rapid motion or complex scene geometry. 
More recent geometry-centric interpolation approaches primarily focus on sparse LiDAR point cloud sequences, limiting their applicability to dense per-pixel depth maps \cite{lu2021pointinet,zeng2022idea,zheng2023neuralpci,zhang2024fastpci}.
Furthermore, to the best of our knowledge, there is still no unified framework that jointly performs RGB and depth interpolation, leaving the joint interpolation of appearance and geometry largely unexplored. 
However, RGB provides rich appearance cues, while depth encodes geometric structure, and the two modalities are inherently complementary for scene understanding. 

In addition, recent advances in event cameras (or dynamic vision sensors) \cite{gallego2020event} have emerged as a promising sensing modality for capturing high-temporal-resolution visual dynamics between frames. 
Unlike conventional frame-based cameras that sample intensity images at fixed rates, event cameras asynchronously record per-pixel brightness changes whenever the logarithmic intensity variation exceeds a predefined threshold.
This asynchronous sensing mechanism produces a sparse yet temporally dense stream of events, providing low-latency and fine-grained motion evidence that is unavailable in standard imaging sensors.
Such micro-temporal cues are particularly valuable for frame interpolation, where the recovery of intermediate dynamics is fundamentally under-constrained from sparse boundary observations alone.
Recent advances in event-guided RGB video interpolation have demonstrated that incorporating event information can substantially improve the robustness of motion estimation and enhance reconstruction quality, especially under fast, abrupt, or irregular motion \cite{Tulyakov21cvpr,tulyakov2021time,ma2024timelens,liu2025timetracker,kim2023event}. 
However, the potential of event-based cues for depth interpolation remains unexplored, and consequently has not been investigated in the context of RGB-D interpolation.


To address these limitations, we propose \systemname{}, a unified RGB-D video frame interpolation framework that jointly leverages RGB, depth and events. 
To the best of our knowledge, \systemname{} is the first unified framework that jointly models RGB and depth interpolation while incorporating RGB, depth and event modalities within a single architecture, enabling the tasks to benefit from their complementary properties and improving computational efficiency through shared representations.
Our framework effectively exploits the distinct roles of the three modalities: RGB provides rich appearance information, depth supplies reliable geometric structure for geometry-consistent interpolation, and events contribute temporally dense intermediate motion cues between boundary frames.
Specifically, our method takes RGB frames, depth maps and event streams as inputs, and processes them through a multimodal feature extraction and fusion module.
Based on the fused representations, we estimate optical flow and synthesize temporally denser RGB-D frames.

Additionally, to address the scarcity of RGB-D-Event data, we construct a new dataset comprising synchronized RGB, dense depth maps, and event streams, which provides a valuable benchmark for training and evaluating tri-modal video interpolation methods. We further conduct extensive experiments to validate the effectiveness of \systemname{} and demonstrate its advantages over existing methods.
In summary, our work makes the following contributions:
\begin{itemize}
    \item We propose \systemname{}, a unified framework that jointly models RGB and depth interpolation within a single architecture, enabling the two tasks to benefit from complementary appearance and geometric cues.
    \item We introduce a tri-modal framework that jointly leverages RGB, depth, and events, effectively integrating appearance, geometry, and dense temporal motion for more accurate and geometry-consistent interpolation.
    \item We construct a new dataset containing synchronized RGB frames, dense depth maps and event streams to facilitate training and evaluation for RGB-D video interpolation and to provide a new benchmark for future research.
\end{itemize}


\section{Related Works}
\subsection{RGB Video Frame Interpolation}
RGB-only video frame interpolation has established strong flow-and-synthesis baselines. 
Super SloMo \cite{jiang2018super} jointly learns bidirectional flow and per-pixel visibility for moderate motion and occlusion. 
PerVFI \cite{wu2024perception} improves perceptual quality through asymmetric blending and a flow-based generator.
RIFE \cite{huang2022real} improves efficiency by directly predicting intermediate flow with privileged distillation. 
However, these methods remain limited by the lack of observations between boundary frames. In highly dynamic or non-stationary scenes, their assumptions on simple motion and appearance constancy often break down, causing motion drift, blur, and ghosting artifacts.

To mitigate this limitation, event-guided approaches introduce micro-temporal cues to compensate for the temporal blind spot of frame-based sensing.
TimeLens \cite{tulyakov2021time} combines events and frames to unify flow warping and pixel synthesis for fast and nonlinear motion. 
TimeLens++ \cite{Tulyakov21cvpr} further models nonlinear dynamics with a one-shot continuous flow representation and multi-scale feature fusion. 
TimeLens-XL \cite{ma2024timelens} handles large displacements through recursive small-motion updates for lightweight inference. 
Similarly, CBMNet \cite{kim2023event} learns event-image fusion and bidirectional flows end-to-end via interactive attention, and TimeTracker \cite{liu2025timetracker} reformulates interpolation as an event-guided local trajectory tracking task to better deal with nonlinear motion. 

However, these methods still focus mainly on RGB interpolation and lack explicit geometric constraints.
Without reliable depth, they remain prone to visibility ordering errors and ambiguous correspondences, especially under large motion or complex occlusion.
Although introducing geometry into a single RGB frame can provide partial structural cues \cite{bao2019depth}, it lacks metric depth and remains unreliable in temporally discontinuous scenes. 

\subsection{Depth Map and Point Cloud Interpolation  }
With the growing interest in 3D vision, dense depth map interpolation has also been studied in several prior works, but remains far less explored than RGB video interpolation.
Early works addressed temporal upsampling through range-color fusion~\cite{dolson2010upsampling}, texture-guided interpolation~\cite{zhang2014texture}, and hybrid-camera reconstruction with scene flow and intermediate depth estimation~\cite{yuan2018temporal}. 
However, these methods are mostly analytical pipelines based on hand-crafted priors, often requiring special hardware or 2D motion cues, and tend to break down under rapid motion, complex geometry, and depth discontinuities.

Meanwhile, a substantial body of work focuses on sparse LiDAR point clouds instead.
PointINet~\cite{lu2021pointinet} estimates bidirectional 3D scene flow and fuses warped point sets with attention.
IDEA-Net~\cite{zeng2022idea} begins with linear interpolation and further compensates for higher-order motion using graph-based features.
NeuralPCI~\cite{zheng2023neuralpci} models multi-frame dynamics with a 4D spatio-temporal field for motion-aware synthesis.
FastPCI~\cite{zhang2024fastpci} balances accuracy and speed using pyramidal features and Transformer-style global reasoning. 
However, these methods are designed for sparse, unordered point clouds with sensor-specific sampling patterns and heavy spatial downsampling,
making them unsuitable for dense per-pixel depth interpolation, where sharp boundaries and fine occlusion structures must be preserved.

In practice, RGB-only interpolation suffers from ambiguous motion and visibility reasoning due to the lack of explicit geometry, while depth interpolation remains difficult near depth discontinuities, disocclusions, and complex geometric variation.
More fundamentally, both settings lack temporally dense observations between boundary frames. 
These limitations motivate a unified framework that jointly exploits RGB appearance, depth geometry, and event-based temporal cues for consistent RGB-D interpolation.

\section{Method}
\label{sec:method}
  \begin{figure*}[!tp]
    \centering
    \includegraphics[width=\textwidth]{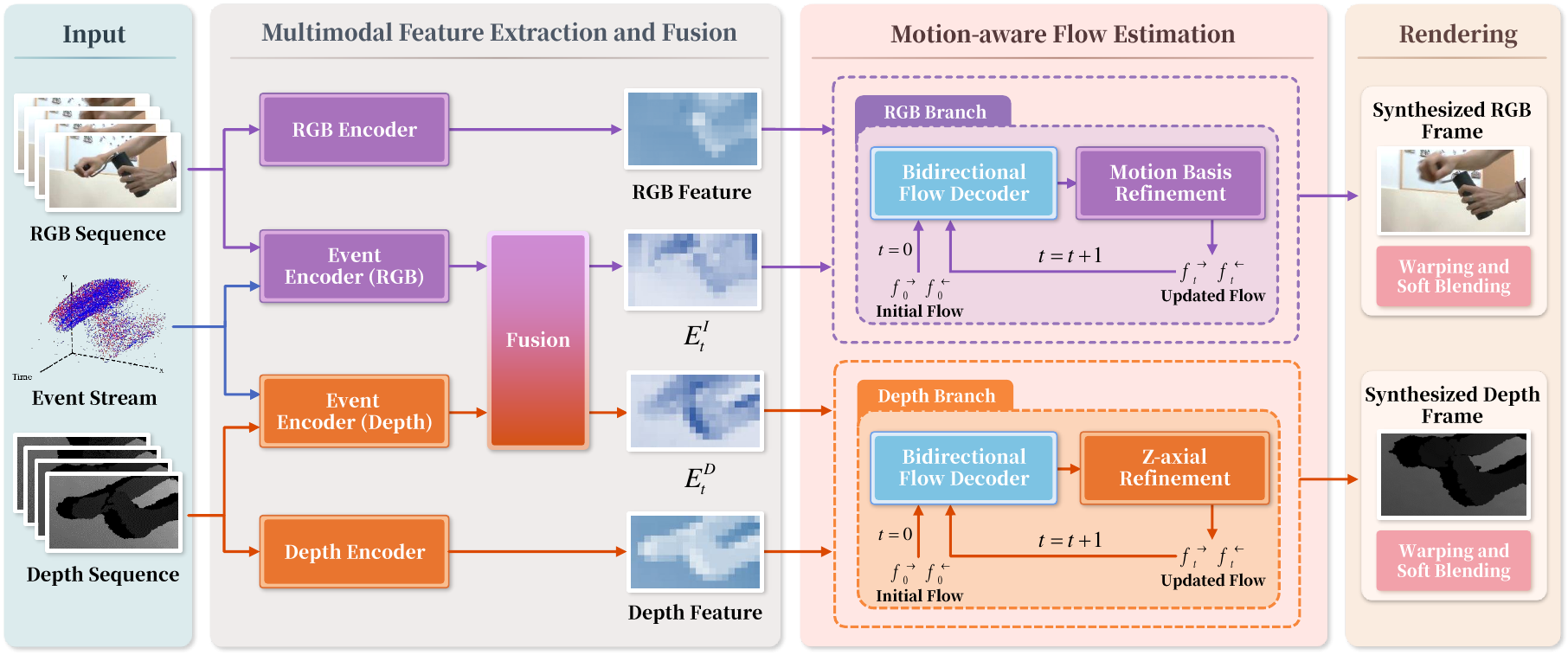}
    \caption{The pipeline of \systemname{}: it consists of three major modules as shaded in different colors along with the input modalities. From left to right, the model takes consecutive RGB frames, depth maps, and the event stream as input. Modality-specific encoders and a tri-modal fusion module first extract and exchange complementary features across RGB, depth, and events. The motion-aware flow estimation module then predicts bidirectional flows in two branches, where motion basis refinement is used for RGB and Z-axial refinement is used for depth. Finally, forward/backward warping and soft blending synthesize the target RGB-D frame.}
    \label{fig:pipeline}
  \end{figure*}
\subsection{Overview}
As illustrated in Fig.~\ref{fig:pipeline}, \systemname{} addresses RGB-D video interpolation by taking consecutive RGB frames, the corresponding depth maps, and the event streams between them as input, and synthesizing the intermediate RGB frame and depth map at the target time.
\systemname{} consists of three major modules: multimodal feature extraction and fusion, motion-aware flow estimation, and rendering.
The first module extracts and exchanges complementary features across modalities using modality-specific encoders and a tri-modal fusion network. 
Based on the fused representations, the second module then estimates bidirectional optical flows for the RGB and depth branches, where depth-guided motion bases are introduced in the RGB branch to provide geometry-aware motion priors, and Z-axial refinement is applied in the depth branch to better capture depth-specific structural variations.
Finally, the rendering module synthesizes interpolated RGB-D frames through forward and backward warping, followed by a soft-blending mechanism that adaptively blends the rewarped results to achieve smooth and low-artifact transitions. 

\subsection{Multimodal Feature Extraction and Fusion}
\label{section:3.1}

RGB, depth, and event data provide complementary observations of the same scene: RGB frames provide rich appearance and texture cues, depth maps contribute explicit geometric structure, and event streams capture per-pixel brightness changes asynchronously at microsecond temporal resolution.
To make effective use of these heterogeneous but complementary signals, we design a multimodal feature extraction and fusion module that extracts modality-specific representations while establishing effective interactions across modalities.

\subsubsection{Feature Extraction}
We first prepare the multimodal inputs for \systemname{}.
Let $F_t^{m}$ denote the observation of modality $m \in \{I, D\}$ at time $t \in \{0,1\}$, where $t=0$ and $t=1$ correspond to the two boundary frames.
Specifically, $F_t^{I} \in \mathbb{R}^{H \times W \times 3}$ represents RGB frame, and $F_t^{D} \in \mathbb{R}^{H \times W}$ denotes the corresponding depth map.
The raw event stream observed between $F_0^{m}$ and $F_1^{m}$ is accumulated and converted into a voxelized representation $\mathcal{E} \in \mathbb{R}^{C \times H \times W}$ with $C$ temporal bins, following \cite{gehrig2019end}.
This representation yields a dense spatio-temporal tensor for feature extraction while preserving temporal variation.

To extract complementary appearance, geometry, and motion representations, \systemname{} employs four dedicated encoders: an RGB encoder, a depth encoder, and two event encoders conditioned on RGB and depth, respectively. The RGB encoder takes the boundary frames $[F_0^{I},\, F_1^{I}]$ as input and extracts multi-scale appearance features using an FPN-style architecture \cite{lin2017feature}. 
Before hierarchical encoding, we introduce channel-wise reweighting and an edge-response map to enhance informative appearance cues.
The depth encoder takes the boundary depth maps $[F_0^{D},\, F_1^{D}]$ as input and extracts multi-scale geometric features using the same FPN-style architecture. 
To strengthen structural representation, we incorporate geometry-aware weighting combined with gradient and continuity responses prior to multi-scale encoding.

The RGB-conditioned event encoder takes $[F_0^{I},\,\mathcal{E},\,F_1^{I}]$ as input and produces multi-scale, time-indexed embeddings $E_{t,s}^{I}$, where $t$ indexes temporal bins and $s$ denotes the feature scale. Specifically, the boundary frames $[F_0^{I},\, F_1^{I}]$ and voxelized events $\mathcal{E}$ are first jointly encoded, then reorganized into slice-wise representations with a fixed channel width, and further compressed through two spatial downsampling stages to obtain compact features $E_{t,s}^{I}$. 
Similarly, the depth-conditioned event encoder follows the same pipeline, taking $[F_0^{D},\,\mathcal{E},\, F_1^{D}]$ as input and producing $E_{t,s}^{D}$. This dual-encoder design not only enables effective extraction of event features but also prepares them to serve as a bridge for subsequent cross-modal fusion between appearance and geometry.

\subsubsection{Feature Fusion}

The extracted appearance features encode rich visual details but do not explicitly capture structural discontinuities, while geometric features provide explicit structural cues but suffer from limited resolution, missing regions, and local aliasing artifacts. 
Therefore, we seek to fuse the two representations to leverage their complementary strengths. 
However, our experiments (Sec.~\ref{ablationstudy}) reveal that naive fusion can be detrimental: RGB details may leak across depth discontinuities, while depth defects (e.g., low resolution, holes, and aliasing) can in turn corrupt appearance features.

To address these limitations of direct fusion, we exploit the event domain as a more suitable space for cross-modal interaction.
Event streams capture motion-induced intensity changes with high temporal precision, and their responses often align with both RGB boundaries and depth discontinuities of moving objects. 
This makes the event domain a more suitable space for exchanging complementary information between the RGB and depth branches than direct feature fusion. 
Motivated by this, we propose a novel fusion strategy that leverages event features as an intermediate bridge, enabling effective cross-modal interaction between appearance and geometry.

At each feature scale $s$, the RGB- and depth-conditioned event features, $E_{t,s}^{I}$ and $E_{t,s}^{D}$, are first projected into a shared embedding space via learned pointwise alignment layers. We then fuse the aligned features using a coordinate-attention interaction module, which captures not only cross-channel dependencies but also direction-aware and position-sensitive information \cite{hou2021coordinate}. 
This preserves positional information while enabling spatially selective feature modulation, making it suitable for event-guided fusion, where informative responses are typically localized around moving boundaries and depth discontinuities.

In practice, the fusion follows the axis-wise reweighting strategy of \cite{hu2024cross}, i.e., aligning channels via learned pointwise linear maps to form: 
\begin{equation}
\begin{aligned}
        X^{\mathrm{I}}_{t,s}=\mathrm{Align}_{\mathrm{I}}(E^{\mathrm{I}}_{t,s}),\quad
X^{\mathrm{D}}_{t,s}=\mathrm{Align}_{\mathrm{D}}(E^{\mathrm{D}}_{t,s}),\quad\\
Z_{t,s}=[\,X^{\mathrm{I}}_{t,s},\,X^{\mathrm{D}}_{t,s}\,].\qquad\qquad\qquad
\end{aligned}
\end{equation}
From the fused feature $Z_{t,s}$, height-wise and width-wise descriptors are used to generate the axis-aware gates $A_h$ and $A_w$.
A spatial saliency gate $S_{t,s}$ is further computed from the channel-wise maximum response, so that the fused representation can focus on spatially informative regions while suppressing inconsistent activations.
The gating and fusion process is expressed as  
\begin{equation}
\begin{aligned}
S_{t,s} &= \sigma\!\Big(\mathrm{Conv}_{k\times k}\big(\max\nolimits_{c}\big(Z_{t,s}\odot A_h\odot A_w\big)\big)\Big),\\
\hat{Z}_{t,s} &= \mathrm{Conv}_{1\times1}\!\big((Z_{t,s}\odot A_h\odot A_w)\odot S_{t,s}\big).
\end{aligned}
\end{equation}
The fused cross-modal message is then injected back into the two event streams symmetrically~\cite{zhang2023cmx} at every temporal bin $t$:
\begin{equation}
\hat{E}^{\mathrm{I}}_{t,s}=E^{\mathrm{I}}_{t,s}+ \mathrm{Align}_{\mathrm{I}}(\hat{Z}_{t,s}),\qquad
\hat{E}^{\mathrm{D}}_{t,s}=E^{\mathrm{D}}_{t,s}+ \mathrm{Align}_{\mathrm{D}}(\hat{Z}_{t,s}).
\end{equation}
In this way, complementary information is exchanged through temporally aligned event responses, which reduces direct contamination between RGB and depth features while preserving the modality-specific characteristics of each branch.

\subsection{Motion-aware Flow Estimation}
\label{section:3.2}
After multimodal feature extraction and fusion, \systemname{} employs a motion-aware flow estimation module to model scene dynamics for frame interpolation. 
As shown in the third column of Fig.~\ref{fig:pipeline}, this module adopts a two-branch architecture for RGB and depth, where each branch employs a phase-wise iterative loop to estimate bidirectional optical flow \cite{teed2020raft,gehrig2021raft}.  
To encourage each branch to focus on its respective role, a motion basis refinement is introduced in the RGB branch to enable appearance-driven motion estimation, while a Z-axial refinement is adopted in the depth branch to enforce geometry-consistent motion within the same iterative framework.
Concretely, let $t \in \{0, \dots, T-1\}$ index the discrete temporal steps, where $T$ denotes the total number of interpolation phases, and $\tau_t = (t+1)/T \in (0,1]$ denotes the corresponding normalized interpolation phase between the two boundary frames.
At each phase $\tau_t$, cumulative flows from both temporal directions are updated in parallel. 
The forward flow is initialized as $\mathbf{f}^{\rightarrow}_0 = 0$ from the starting boundary frame $F_0^{I,D}$, while the backward flow is initialized as $\mathbf{f}^{\leftarrow}_0 = 0$ from the ending boundary frame $F_1^{I,D}$. 
For a pixel location $x$, the forward flow is updated as
\begin{equation}
\mathbf{f}^{\rightarrow}_{t+1}(x)
=\mathbf{f}^{\rightarrow}_{t}(x)
+\Delta\mathbf{f}^{\rightarrow}_t\!\big(x+\mathbf{f}^{\rightarrow}_{t}(x)\big),
\end{equation}
driven by the event slice $E_t^{I,D}$, while the backward flow follows
\begin{equation}
\mathbf{f}^{\leftarrow}_{t+1}(x)
=\mathbf{f}^{\leftarrow}_{t}(x)
+\Delta\mathbf{f}^{\leftarrow}_t\!\big(x+\mathbf{f}^{\leftarrow}_{t}(x)\big),
\end{equation}
driven by the time-reversed event slice $E_{T-1-t}^{I,D}$. 
The resulting flows $\mathbf{f}^{\rightarrow}_{t}$ and $\mathbf{f}^{\leftarrow}_{t}$ represent cumulative displacements from the two boundary frames to the intermediate phase $\tau_t$, respectively.

Given these flows, the two boundary frames are warped toward the current phase to produce forward and backward candidates, which serve as both reconstruction hypotheses and inputs for subsequent updates. 
The estimation proceeds sequentially from $\tau_t$ to $\tau_{t+1}$ until all intermediate frames are generated. 
During training, supervision is applied at each phase, providing dense optimization signals that stabilize temporal prediction.

At each iteration of the phase-wise bidirectional estimation loop, the RGB and depth branches adopt distinct refinement strategies tailored to their respective roles: motion basis refinement for the RGB branch and Z-axial refinement for the depth branch. 
We next detail these two refinement strategies. 

\subsubsection{Motion Basis refinement}
Under the pinhole camera model and assuming small inter-frame motion, the instantaneous image velocity at normalized image coordinates can be decomposed into a translational component $\mathbf{v}\in\mathbb{R}^3$ and a rotational component $\boldsymbol{\omega}\in\mathbb{R}^3$. 
A key property is that the translational term depends on inverse depth $\zeta = 1/Z$, whereas the rotational term is independent of depth \cite{hartley2003multiple,longuet1980interpretation,bruss1983passive}. 
This yields a compact and physically grounded motion parameterization.
Under this formulation, optical flow can be represented as a linear combination of fixed basis vector fields, $\{\Delta T_x, \Delta T_y, \Delta T_z, \Delta R_x, \Delta R_y, \Delta R_z\}$ 
defined in pixel coordinates and parameterized by the camera intrinsics \cite{heeger1992subspace}. 
Following Poggi and Tosi \cite{poggi2025flowseek}, we assume $f_x=f_y$, and split $\Delta R_x$ and $\Delta R_y$ into two sub-bases each, resulting in an eight-basis representation. 
We denote this predefined motion basis set as 
$\mathcal{M}=\{\mathbf{b}_k\}_{k=1}^{K}$ with $K=8$, whose elements are fixed and encode the geometric structure of rigid motion.

Given the endpoint depth maps $F_0^D$ and $F_1^D$, we use the generator introduced by Poggi and Tosi \cite{poggi2025flowseek} to estimate endpoint-specific motion priors. 
Specifically, the generator predicts soft, spatially varying coefficients to combine the fixed basis set $\mathcal{M}$, yielding two basis-induced motion fields 
$\widetilde{\mathcal{M}}^{(0)}, \widetilde{\mathcal{M}}^{(1)} \in \mathbb{R}^{H \times W \times 2}$. 
Here, $\widetilde{\mathcal{M}}$ denotes the resulting motion prior obtained by weighting $\mathcal{M}$ according to depth.

To handle general motion, we assume that motion is locally rigid in space and evolves smoothly over time.  
For an intermediate time $\tau_t \in [0,1]$, we obtain the corresponding motion prior by linearly interpolating the endpoint priors:
\begin{equation}
\widetilde{\mathcal{M}}^{(\tau_t)}
=
(1 - \tau_t)\,\widetilde{\mathcal{M}}^{(0)}
+
\tau_t\,\widetilde{\mathcal{M}}^{(1)}.
\end{equation}
This yields a continuous trajectory in the motion basis subspace and provides an explicit temporal constraint for intermediate flow estimation.

Conditioned on $\widetilde{\mathcal{M}}^{(\tau_t)}$, we employ a motion-basis-guided predictor $\mathcal{H}_{\mathrm{mb}}$, implemented as a lightweight convolutional module, to predict a residual flow 
$\Delta\mathbf{f}_{\mathrm{mb},t}$ together with a confidence map $\mathbf{m}_{\mathrm{mb},t}\in[0,1]$:
\begin{equation}
(\Delta\mathbf{f}_{\mathrm{mb},t},\,\mathbf{m}_{\mathrm{mb},t})
= 
\mathcal{H}_{\mathrm{mb}}\!\big(\widetilde{\mathcal{M}}^{(\tau_t)}\big).
\end{equation}
Here, the residual term provides a motion basis correction to the current flow estimate, while the confidence map indicates the pixel-wise reliability of the motion prior. 
We then apply the correction in a gated manner:
\begin{equation}
\mathbf{f}_t^{+}
=
\mathbf{f}_t
+
\sigma\!\big(\mathbf{m}_{\mathrm{mb},t}\big)
\odot
\Delta\mathbf{f}_{\mathrm{mb},t}.
\end{equation}
In this way, the model applies stronger updates where the motion prior is reliable and suppresses them where the prior is less trustworthy.
Note that the motion basis refinement is used only in the RGB branch.
This is because RGB interpolation benefits more from explicit geometric guidance for dense appearance motion estimation. By contrast, depth is more affected by residual errors along the viewing direction, so it is refined with a dedicated Z-axial module instead.

\subsubsection{Z-axial refinement}
2D optical flow captures only image-plane motion and is therefore insensitive to axial motion along the viewing direction.
This limitation is particularly evident under large parallax and occlusion, where depth errors are often caused by residual optical-axis displacement \cite{vedula1999three,ma2004invitation}.

To address this issue, we introduce a Z-axial refinement module in the depth branch to adjust only the z-axial component while preserving the in-plane estimate.
Specifically, at phase $t$, the Z-axial refinement module takes three inputs: the current 2D flow features $\mathbf{f}_t$, depth features $\hat{F^D_t}$, and temporally aggregated event features $\mathcal{A}(E_t)$. 
These inputs are chosen to provide complementary constraints for axial correction. 
The flow features encode the current in-plane motion estimate and thus serve as the reference state to be corrected. 
The depth features provide scene geometry that determines how displacement along the viewing direction affects reprojection. 
The aggregated event features supply fine-grained temporal evidence of residual motion, which is particularly useful when axial errors are not fully captured by the current 2D flow estimate. 
Based on these cues, the module predicts an axial residual $\Delta f_t^{(z)}$ together with a confidence map $\mu_t^{(z)}$:
\begin{equation}
\big(\Delta f^{(z)}_{t},\,\mu^{(z)}_{t}\big)
=\mathcal{Z}\!\Big(
\mathbf{f}_{t},\hat{F^D_t},
\mathcal{A}(E_{t})
\Big).
\end{equation}
Let $w_t^{\mathrm{base}}$ denote the initial warp obtained from the current flow estimate before applying the z-axial refinement, the predicted axial correction is injected into this warp as:
\begin{equation}
w_t^{+} = w_t^{\mathrm{base}} + \mu_t^{(z)} \odot \Delta f_{t}^{(z)}.
\end{equation}

In this way, the module selectively strengthens axial correction where the estimated residual is reliable, while suppressing unstable updates in ambiguous regions. 


\subsection{Rendering and Loss Functions}
\label{section:3.3}

\subsubsection{Rendering}
After all prior refinements, the Motion-aware Flow Estimation module outputs the forward/backward flows, their coarse rewarps, and the associated decoder flow features.
For synthesis, we adopt an asymmetric soft-blending strategy inspired by Wu~et~al.~\cite{wu2024perception}.
Rather than simply averaging the two coarse rewarps, the module predicts a spatially varying soft visibility map to adaptively fuse the bidirectional warped candidates. 
Specifically, the binary mask is first encoded by a convolutional layer stack to produce mask features. 
In parallel, the two warped candidates are concatenated and processed by a lightweight attention module to generate feature-dependent modulation weights.
The modulation weights then refine the mask features to produce the final soft blending mask, which enables pixel-wise fusion according to local reliability.
This design allows the fusion process to adapt to regions where the two directional warps differ, such as occlusion areas and motion boundaries. 
The resulting fused image is taken as the final interpolated result at the target time step. 

\subsubsection{Loss functions}
We adopt branch-specific supervision for appearance and geometry.  
First, since both RGB and depth interpolation require stable per-pixel regression under large local variations, we employ the Charbonnier loss as the primary reconstruction objective for both branches \cite{sun2014quantitative,barron2019general}:

\begin{equation}
\mathcal{L}_{\text{Char}}(\hat F_t, F_t)
=\frac{1}{N}\sum_{i=1}^{N}\sqrt{\big(\hat F_{t,i}-F_{t,i}\big)^2+\varepsilon},
\label{eq:char_rgb}
\end{equation}
where $\hat F_t$ denotes the predicted frame, $F_t$ represents the ground truth, $i$ denotes the index of a valid pixel, $N$ denotes the total number of valid pixels and $\varepsilon>0$ is a small constant introduced for numerical stability. 

For the RGB branch, we additionally employ an LPIPS term \cite{zhang2018unreasonable} to preserve perceptual image quality, resulting in the final loss for RGB interpolation:
\begin{equation}
\begin{aligned}
\mathcal{L}_{\text{RGB}}
=\sum_{t=0}^{T-1}\Big[\mathcal{L}_{\text{Char}}(\hat {F_t^I},F_t^I)
+\lambda_{\text{p}}\ \mathrm{LPIPS}(\hat {F_t^I},F_t^I)\Big].
\end{aligned}
\label{eq:loss_rgb}
\end{equation}

For the depth branch, we augment the Charbonnier term with a first-order gradient consistency loss \cite{bhat2021adabins,godard2017unsupervised,godard2019digging} to preserve depth boundaries and coherent local gradient structures, which penalizes discrepancies in the horizontal and vertical depth derivatives:
\begin{equation}
\mathcal{L}_{\nabla}(\hat {F_t^D},F_t^D)
=\big\|\nabla_x \hat {F_t^D}-\nabla_x F_t^D\big\|_{1}
+\big\|\nabla_y \hat {F_t^D}-\nabla_y F_t^D\big\|_{1}.
\label{eq:grad_loss}
\end{equation}
Based on this formulation, the final loss for depth interpolation is defined as
\begin{equation}
\begin{aligned}
\mathcal{L}_{\text{Depth}}
=\sum_{t=0}^{T-1}\Big[\mathcal{L}_{\text{Char}}(\hat {F_t^D},F_t^D)
+\lambda_{\nabla}\ \mathcal{L}_{\nabla}(\hat {F_t^D},F_t^D)\Big].
\label{eq:loss_depth}
\end{aligned}
\end{equation}

Finally, the two branch losses are combined as
\begin{equation}
\mathcal{L}_{\text{total}}
= \alpha\,\mathcal{L}_{\text{RGB}}
+\beta\,\mathcal{L}_{\text{Depth}} ,
\label{eq:loss_total}
\end{equation}
where $\alpha$ and $\beta$ are fixed weighting coefficients that balance the contributions of the RGB and depth losses, respectively.

\section{Datasets}
\label{data}

\systemname{} simultaneously relies on RGB, depth, and event modalities, yet existing datasets lack a dedicated tri-modal benchmark. 
To enable systematic evaluation, we construct two datasets: one derived from processed VECtor data \cite{gao2022vector}, and our newly collected SyncRDE-60 dataset.


\subsection{VECtor}
\label{sec:vector_preprocess}

We use a small-scale subset of VECtor \cite{gao2022vector}, which provides dense depth annotations, to construct a geometrically consistent RGB–D–event benchmark. 
\begin{figure}[tbp]
  \centering
  \includegraphics[width=0.8\linewidth]{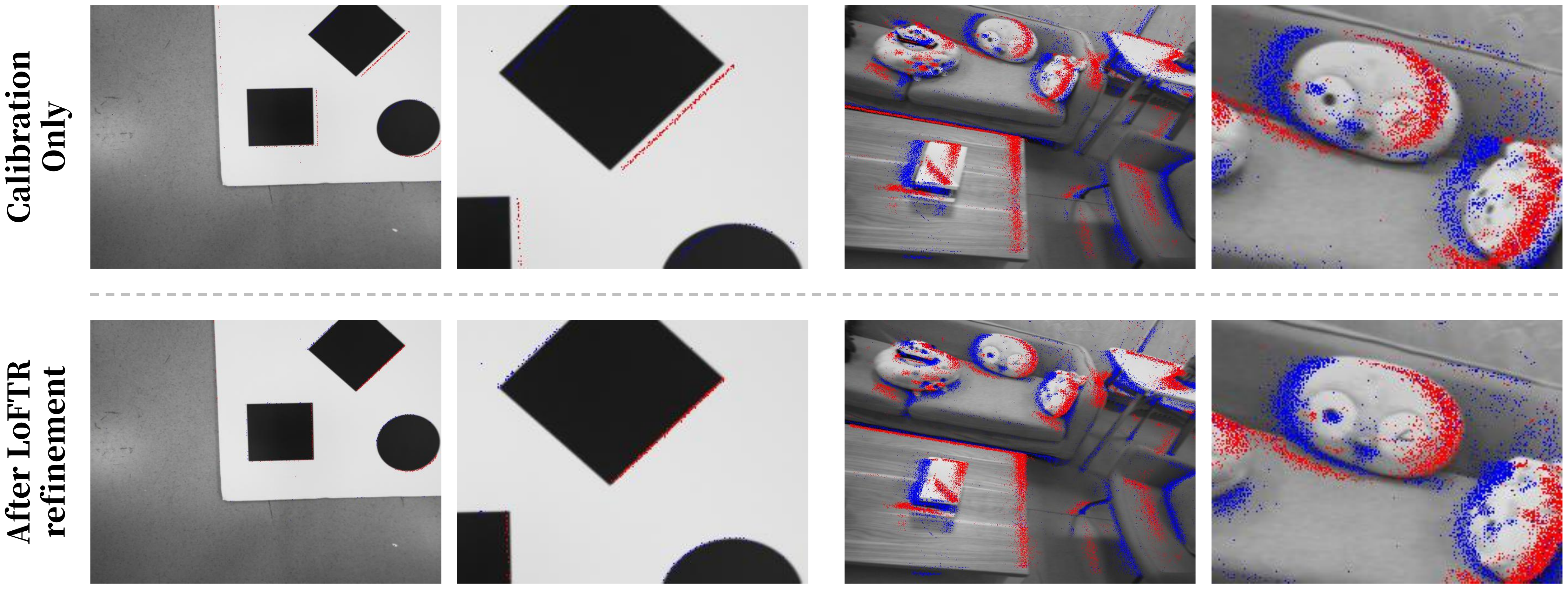}
  \caption{Qualitative comparison of alignment before and after LoFTR-based residual refinement.}
  \label{fig:loftr}
\end{figure}

The raw sequences are processed in two stages. 
In the first stage, RGB frames are undistorted, and depth maps are warped to a common RGB reference frame using the provided calibration parameters. 
Each event is first undistorted in the event camera domain, then associated with its temporally nearest depth frame to obtain a depth value via bilinear interpolation, and finally reprojected to the target RGB image plane using the calibrated extrinsics.

In the second stage, we further correct residual RGB–event misalignment, as calibration-based preprocessing alone may be insufficient in close-range scenes, as also observed in TimeLens \cite{tulyakov2021time}. 
To address this, we refine the alignment via LoFTR matching followed by RANSAC-based residual correction. 
Specifically, for each sequence, we sample a set of reference frames and accumulate nearby events around each timestamp to construct event images. 
We then establish cross-modal correspondences between the event images and the edge maps of the corresponding RGB frames using LoFTR \cite{sun2021loftr}. 
After confidence filtering, residual geometric corrections are estimated via RANSAC \cite{fischler1981random}, averaged at the sequence level, and applied to the entire event stream. We visualize the alignment quality before and after refinement in Fig.~\ref{fig:loftr}.
The visual comparison shows 
that, after refinement, the alignment exhibits substantially improved consistency between event structures and RGB boundaries.

\subsection{SyncRDE-60} 
\begin{figure}[tb]
  \centering
  \includegraphics[width=0.8\linewidth]{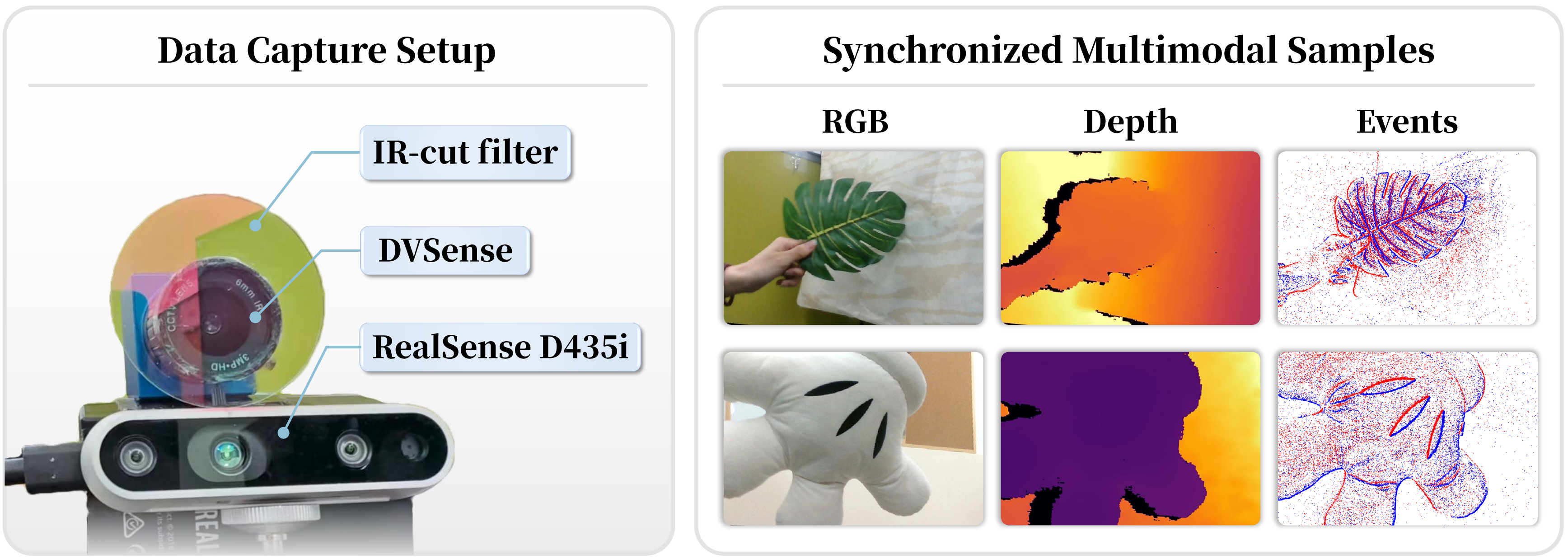}
  \caption{Left: our multimodal data capture platform, consisting of a DVSense event camera with an IR-cut filter mounted on an Intel RealSense D435i. Right: representative samples from the captured dataset, including RGB images, depth maps, and event visualizations. Red and blue denote positive and negative event polarities, respectively. All modalities are spatially calibrated and registered to a common reference frame.}
  \label{fig:equip}
\end{figure}

We also construct a new dataset, SyncRDE-60, by collecting and curating 60 sequences of synchronized RGB-D videos and event streams.
For data acquisition, we build a capture platform as shown in Fig.~\ref{fig:equip}. 
RGB-D data are captured using an Intel RealSense D435i at a resolution of $640\times360$ and 30 fps, while event data are recorded using a DVSense sensor at a resolution of $1280\times720$ \cite{realsense,dvsense}. 
Both devices are connected to the same host computer, ensuring that all measurements share a common system clock and can be temporally aligned within a unified acquisition framework.

For spatial alignment, we adopt a calibration-based geometric pipeline. 
RGB and depth are first registered using the factory calibration and built-in alignment of the Intel RealSense D435i. 
Each sensor is independently calibrated to obtain its intrinsics and distortion coefficients.
Following \cite{dubeau2020rgb}, we estimate the extrinsic transformation from the RGB-D reference frame to the event camera using a jointly observed checkerboard. 
Checkerboard corners are detected in the RGB frames and event streams, and the aligned depth maps are used to lift the RGB corners to 3D in the RGB-D reference frame. The extrinsic transformation is then solved by PnP \cite{lepetit2009ep} from the recovered 3D points, the event-image corners, and the calibrated event intrinsics.
With the estimated RGB-D-to-event extrinsics, we further align the three modalities using the same preprocessing pipeline as that described in Sec.~\ref{sec:vector_preprocess}.

The collected sequences primarily capture object motions driven by human manipulation, along with a subset induced by indirect off-camera actuation to enrich motion diversity. The dataset spans a wide range of object categories, including human hands and forearms, utensils and containers, balls and toys, food items, consumer electronics, stationery, and various household or office objects.
It also contains challenging motion patterns such as non-rigid deformation, frequent hand–object interactions and occlusions, and dynamics ranging from subtle displacements to rapid movements.
More detailed dataset statistics and split information are provided in the supplementary material.




\section{Experiments and Analysis }
\label{sec:exp}

\subsection{Settings and Metrics.}

All experiments are conducted using the PyTorch framework on a single NVIDIA RTX 4090 GPU with a batch size of 2. 
We set the loss weights as $\lambda_{\text{p}}=0.1$ for the RGB perceptual term, $\lambda_{\nabla}=0.2$ for the depth gradient term, and $\alpha=1.0$, $\beta=0.8$ for the RGB and depth branches, respectively. 
The model is optimized using the Adam optimizer \cite{kingma2014adam} with an initial learning rate of $1\times10^{-5}$, followed by a cosine decay schedule to $1\times10^{-7}$.

For evaluation, we assess the RGB branch using PSNR, SSIM and LPIPS \cite{zhang2018unreasonable}, which measure pixel-level fidelity, structural consistency around edges and occlusions, and perceptual quality, respectively. 
For the depth branch, we report RMSE, AbsRel, and $\delta_{1.25}$. RMSE quantifies absolute geometric error, AbsRel measures relative accuracy across varying depth ranges, and $\delta_{1.25}$ evaluates the proportion of pixels whose predicted depth falls within a tight tolerance of the ground truth.
Together, these metrics provide a comprehensive evaluation of both photometric and geometric performance in RGB–D video interpolation.


 \begin{figure*}[!tp]
    \centering
    \includegraphics[width=0.85\textwidth]{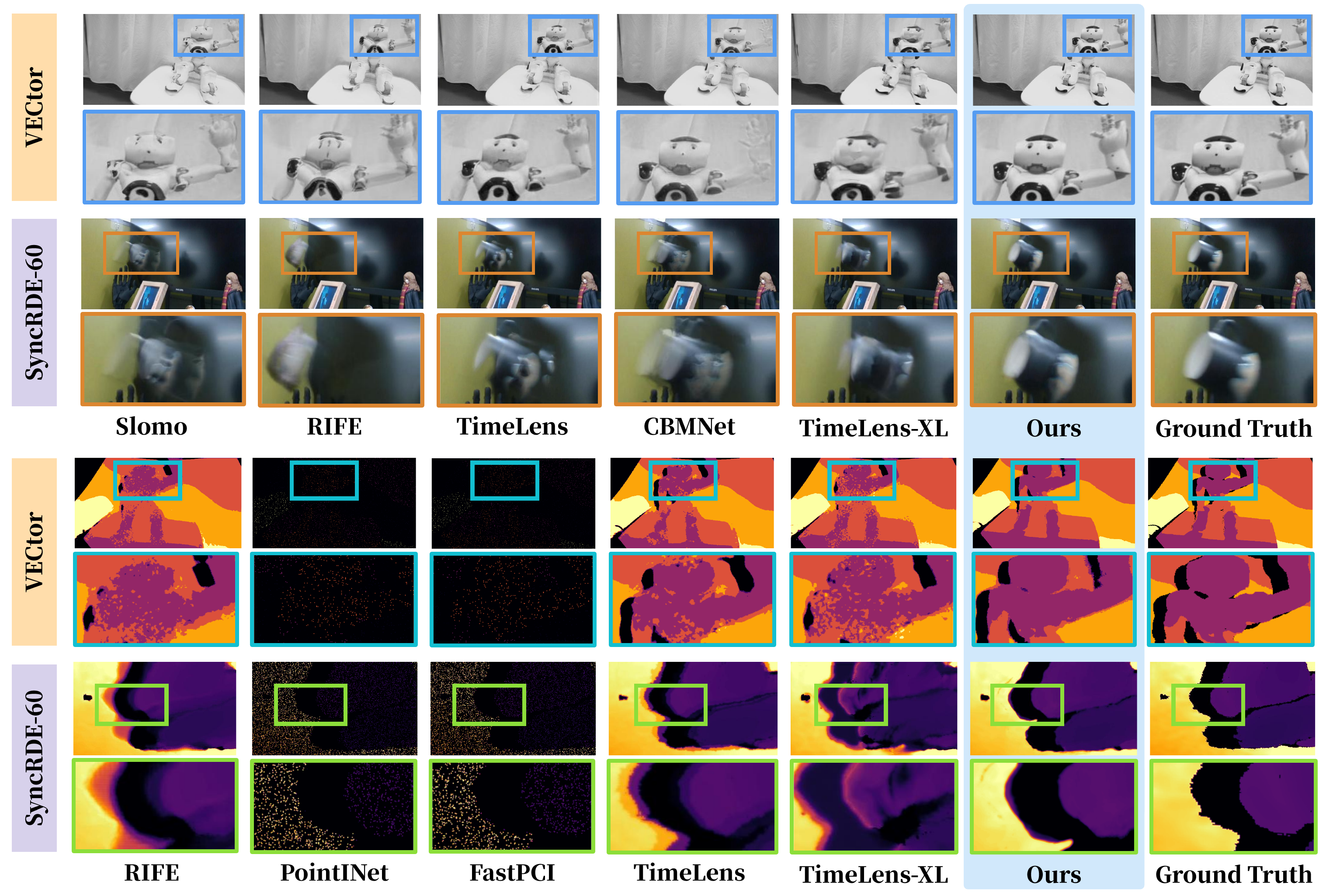}
    \caption{Qualitative comparison on VECtor and SyncRDE-60 at $\times4$. 
    Examples of interpolated RGB frames (upper half) and corresponding depth frames (lower half) from different methods are compared with ground truth in different columns. The partial regions are zoomed in to show detailed interpolation quality. }
    \label{fig:result_rgb}
  \end{figure*}

\subsection{Comparison with Different Methods}
In the absence of established event-guided RGB-D interpolation baselines, we design the evaluation protocol along two complementary tracks to ensure that the comparisons remain fair and informative.
The first track measures the appearance by comparing \systemname{} with RGB frame-based methods (SuperSloMo \cite{jiang2018super}, RIFE \cite{huang2022real}) and event-guided methods (CBMNet \cite{kim2023event}, TimeLens \cite{tulyakov2021time}, TimeLens-XL \cite{ma2024timelens}). 
The second track measures the geometry by comparing against depth interpolation (RIFE with depth input), point-cloud temporal upsampling (PointINet \cite{lu2021pointinet}, FastPCI \cite{zhang2024fastpci}), and event-guided depth interpolation (TimeLens and TimeLens-XL with depth input). 

All methods are retrained and validated independently for each dataset and temporal upsampling protocol.
For point-cloud interpolation baselines (PointINet and FastPCI), we uniformly downsample each point cloud to 8,192 points during both training and inference. 
This step is necessary due to the inherent computational and memory complexity of point-based processing, and is commonly adopted in prior works. 
Evaluation metrics for these methods are computed only on pixels with valid point projections to ensure fair and meaningful comparisons. 
Following this protocol, we train separate models for VECtor $\times4$, VECtor $\times8$, SyncRDE-60 $\times4$, and SyncRDE-60 $\times8$, where $\times k$ denotes $k$-times temporal upsampling.

\subsubsection{Evaluation results of RGB interpolation}
We present qualitative comparisons in the upper half of Fig.~\ref{fig:result_rgb}. \systemname{} produces consistently sharper results with cleaner edges and fewer ghosting artifacts than both frame-only and event-guided baselines. 
For instance, our method faithfully preserves the mug’s shape and surface texture, without noticeable deformation or overlapping artifacts in the SyncRDE-60 example. 
Frame-only methods suffer from temporal instability, while existing event-guided approaches, although improving temporal stability, still lack geometric constraints and tend to exhibit boundary drift and double edges near motion boundaries.

We also report quantitative comparisons with competing methods in Table~\ref{tab:rgb}. Across both VECtor and SyncRDE-60 datasets, \systemname{} consistently achieves the best PSNR, SSIM, and LPIPS under both $\times4$ and $\times8$ settings, with noticeably larger gains at $\times8$ where the larger temporal gap makes interpolation substantially more challenging for all methods.
These results suggest that our method more effectively suppresses residual misalignment and preserves boundary fidelity, rather than merely producing smoother outputs.



\subsubsection{Evaluation results of depth interpolation}

The qualitative results for depth interpolation are shown in the lower half of Fig. \ref{fig:result_rgb}. A consistent trend is observed: \systemname{} produces cleaner and more continuous depth with crisper object boundaries than frame-only and event-guided baselines. Specifically, for the example from VECtor, our result preserves a more complete and coherent depth structure in the head and upper-body regions, while avoiding the scattered holes, fragmented contours, and patchy depth artifacts visible in competing methods. Frame-only depth interpolation drifts when the temporal gap is large and often shows banding. Point-cloud interpolation suffers from sparse sampling and reprojection errors, often leading to holes, broken contours, and temporal jitter around motion boundaries. Event-based models time transitions well but lack metric geometry, so z-order and boundary continuity fail. 

\begin{table*}[t]
\centering
\caption{Quantitative evaluation results for RGB frame interpolation on VECtor and SyncRDE-60. Inputs: R=RGB, D=Depth, E=Events. The best result for each metric is marked as \textbf{bold} and the second best is \underline{underlined}.}
\label{tab:rgb}
\small
\setlength{\tabcolsep}{2pt}
\renewcommand{\arraystretch}{1.1}
\begin{tabular}{l c | c c c | c c c | c c c | c c c}
\hline
\multicolumn{2}{c|}{} & \multicolumn{6}{c|}{VECtor (RGB)} & \multicolumn{6}{c}{SyncRDE-60 (RGB)} \\
\cline{3-14}
\multicolumn{2}{c|}{} & \multicolumn{3}{c|}{$\times 4$} & \multicolumn{3}{c|}{$\times 8$} & \multicolumn{3}{c|}{$\times 4$} & \multicolumn{3}{c}{$\times 8$} \\
\cline{3-14}
Method & Inputs
& PSNR $\uparrow$ & SSIM $\uparrow$ & LPIPS $\downarrow$
& PSNR $\uparrow$ & SSIM $\uparrow$ & LPIPS $\downarrow$
& PSNR $\uparrow$ & SSIM $\uparrow$ & LPIPS $\downarrow$
& PSNR $\uparrow$ & SSIM $\uparrow$ & LPIPS $\downarrow$ \\
\hline
SuperSloMo \cite{jiang2018super}  & R     & 25.3235 & 0.8389 & 0.0103 & 20.9860 & 0.7697 & 0.0195 & 30.5073 & 0.9403 & 0.0079 & 27.1991 & 0.9489 & 0.0124 \\
RIFE \cite{huang2022real}          & R     & 25.8885 & 0.8547 & 0.0089 & 21.7836 & 0.7878 & 0.0181 & 31.6077 & 0.9447 & 0.0064 & 27.6409 & 0.9508 & 0.0099\\
Timelens \cite{tulyakov2021time}    & R+E   & 31.2513 & 0.9144 & \underline{0.0047} & 26.5351 & 0.8501 & \underline{0.0110} & 32.4868 & 0.9598 & 0.0072 & 28.1587 & 0.9517 & 0.0106 \\
CBMNet \cite{kim2023event}         & R+E     & \underline{32.0829} & \underline{0.9200} & 0.0065 & \underline{27.2596} & \underline{0.9047} & 0.0130 & 33.0671 & 0.9364 & 0.0076 & 29.4400 & 0.9536 & 0.0103 \\
Timelens-XL \cite{ma2024timelens}  & R+E   & 28.0381 & 0.8780 & 0.0058 & 23.3609 & 0.8136 & 0.0133 & \underline{33.7292} & \underline{0.9616} & \underline{0.0058} & \underline{30.3999} & \underline{0.9582} & \underline{0.0087} \\
\rowcolor{blue!10} 
\textbf{UniRED}  & R+D+E & \textbf{33.6368} & \textbf{0.9427} & \textbf{0.0027}
                       & \textbf{28.2768} & \textbf{0.9170} & \textbf{0.0074}
                       & \textbf{35.7833} & \textbf{0.9712} & \textbf{0.0054}
                       & \textbf{31.7508} & \textbf{0.9577} & \textbf{0.0065} \\
\hline
\end{tabular}
\end{table*}

\begin{table*}[t]
\centering
\caption{Quantitative evaluation results for depth frame interpolation on VECtor and SyncRDE-60. Inputs: R=RGB, D=Depth, E=Events, PC=Point Clouds. The best result for each metric is marked as \textbf{bold} and the second best is \underline{underlined}.}
\label{tab:depth}
\small
\setlength{\tabcolsep}{2pt}
\renewcommand{\arraystretch}{1.1}
\begin{tabular}{l c | c c c | c c c | c c c | c c c}
\hline
\multicolumn{2}{c|}{} & \multicolumn{6}{c|}{VECtor (Depth)} & \multicolumn{6}{c}{SyncRDE-60 (Depth)} \\
\cline{3-14}
\multicolumn{2}{c|}{} & \multicolumn{3}{c|}{$\times 4$} & \multicolumn{3}{c|}{$\times 8$} & \multicolumn{3}{c|}{$\times 4$} & \multicolumn{3}{c}{$\times 8$} \\
\cline{3-14}
Method & Inputs
& RMSE $\downarrow$ & AbsRel $\downarrow$ & $\delta_{1.25}$ $\uparrow$
& RMSE $\downarrow$ & AbsRel $\downarrow$ & $\delta_{1.25}$ $\uparrow$
& RMSE $\downarrow$ & AbsRel $\downarrow$ & $\delta_{1.25}$ $\uparrow$
& RMSE $\downarrow$ & AbsRel $\downarrow$ & $\delta_{1.25}$ $\uparrow$ \\
\hline
RIFE \cite{huang2022real}      & D     & 0.0021 & 0.0209 & \underline{0.9919} & \textbf{0.0015} & \underline{0.0223} & 0.9692 & 0.0619 & \underline{0.1542} & \underline{0.9130} & \underline{0.0658} & 0.1842 & \underline{0.8955} \\
PointINet \cite{lu2021pointinet}    & D(PC)     & 0.0044 & 0.0352 & 0.7014 & 0.0049 & 0.0511 & 0.6233 & 0.1978 & 0.6027 & 0.6713 & 0.3074 & 0.8049 & 0.6027 \\
FastPCI \cite{zhang2024fastpci}    & D(PC)     & 0.0041 & 0.0343 & 0.7268 & 0.0046 & 0.0487 & 0.6679 &0.1962 & 0.6954 & 0.6974 & 0.2216 & 0.7429 & 0.6506 \\
TimeLens \cite{tulyakov2021time}    & D+E     & \underline{0.0018} & \underline{0.0149} & 0.9857 & 0.0024 & 0.0285 & \underline{0.9721} & \underline{0.0611} & 0.1623 & 0.9006 & 0.0746 & \underline{0.1729} & 0.8900 \\
TimeLens-XL \cite{ma2024timelens}    & D+E     & \underline{0.0018} & 0.0157 & 0.9847 & 0.0022 & 0.0278 & 0.9771 & 0.0641 & 0.1988 & 0.8531 & 0.0660 &0.2348 & 0.8570 \\
\rowcolor{blue!10} 
\textbf{UniRED}  & R+D+E & \textbf{0.0013} & \textbf{0.0103} & \textbf{0.9972}
                       & \textbf{0.0015} & \textbf{0.0196} & \textbf{0.9846}
                       & \textbf{0.0570} & \textbf{0.1031} & \textbf{0.9346}
                       & \textbf{0.0626} & \textbf{0.1292} & \textbf{0.9125} \\
\hline
\end{tabular}
\end{table*}

The quantitative results in Table~\ref{tab:depth} further validate our qualitative observations.
Across both the VECtor and SyncRDE-60 datasets, \systemname{} consistently outperforms all baselines under both $\times4$ and $\times8$ settings, achieving the best performance in RMSE, AbsRel, and $\delta_{1.25}$. 
These metrics are particularly sensitive to errors near depth boundaries. 
Frame-only approaches accumulate drift as temporal gaps increase, point-cloud reprojection introduces sparse inconsistencies, and event-based timing without metric depth fails to preserve correct occlusion ordering. 
As a result, these methods exhibit higher RMSE and lower $\delta_{1.25}$ around motion boundaries.

\subsection{Ablation Study}
\label{ablationstudy}
To verify the effectiveness of each design in \systemname{}, we conduct ablation studies on the SyncRDE-60 dataset.

\subsubsection{Ablation studies on modalities}
We perform ablations by removing each input modality (RGB, depth, and event stream) to evaluate their impact on performance. 
The results suggest that jointly performing RGB and depth interpolation is beneficial to both tasks, as the two modalities can mutually reinforce each other through their complementary appearance and geometric cues.
In addition, the event streams play a central role in connecting the RGB and depth branches: once removed, performance drops markedly in both modalities. This observation indicates that event cues are not merely auxiliary inputs, but provide essential temporal guidance for maintaining cross-modal consistency under dynamic motion. 

\subsubsection{Ablation studies on naive tri-modal baseline}
To examine whether the observed gains can be obtained by simply augmenting an existing RGB+Event framework with depth, we implement a naive RGB-D+Events baseline, denoted as TL (RGB-D+E), by concatenating RGB and depth into a 4-channel input and extending TimeLens to process RGB-D frames.
The results in Table~\ref{tab:abl_rgb} and Table~\ref{tab:abl_depth} show that this simple extension remains inferior to \systemname{}. 
Although the naive tri-modal baseline improves depth estimation over the depth+events variant of TimeLens (Table~\ref{tab:depth}), it performs worse in RGB interpolation than the original TimeLens with RGB+events input (Table~\ref{tab:rgb}). 
This observation indicates that simple channel-wise concatenation is insufficient to exploit the complementary properties of the three modalities effectively.



\subsubsection{Ablation studies on fusion strategies}
We compare the proposed fusion strategy in \systemname{} (denoted as \textbf{F4}) with three alternative variants to evaluate its effectiveness and necessity. 
\textbf{F1} removes the fusion module, directly feeding the event-encoder outputs $E_t^I$ and $E_t^D$ into the two downstream branches. 
\textbf{F2} performs naive fusion by concatenating RGB, depth, and event features to produce shared feature maps for subsequent processing. 
\textbf{F3} adopts the same fusion-module architecture as F4, but restricts its inputs to RGB and depth features only. The event stream is still processed through its dedicated pathway, but is excluded from the explicit RGB–depth fusion stage, thereby removing event-guided interaction while retaining appearance–geometry fusion.

\begin{table}[!t]
\centering
\caption{Ablation results on SyncRDE-60 $\times4$ (RGB) with different selections of input modalities and modules. }
\label{tab:abl_rgb}
\footnotesize
\setlength{\tabcolsep}{3pt}
\renewcommand{\arraystretch}{1}
\begin{tabular}{l c c c c c c c c}
\toprule
ID & R & D & E & MB & Fus & PSNR$\uparrow$ & SSIM$\uparrow$ & LPIPS$\downarrow$ \\
\midrule
V0  & \cmark& \xmark& \cmark& \xmark& F1& 33.9002& 0.9538& 0.0059\\
V1  & \cmark& \cmark& \xmark& \cmark& F4& 33.0157& 0.9622& 0.0065\\
V3& \cmark& \cmark& \cmark& \xmark& F4& 34.1672& 0.9601& 0.0069\\
TL (RGB-D+E)& \cmark& \cmark& \cmark& -& -& 31.4074& 0.9558& 0.0079\\
\midrule
V5& \cmark& \cmark& \cmark& \cmark& F1& 28.9904& 0.9482& 0.0106\\
V6& \cmark& \cmark& \cmark& \cmark& F2& 30.7365& 0.9431& 0.0071\\
 V7& \cmark& \cmark& \cmark& \cmark& F3& 32.0072& 0.9589&0.0067\\
\rowcolor{blue!10} 
UniRED& \cmark& \cmark& \cmark& \cmark& F4& 35.7833& 0.9712& 0.0054\\
\bottomrule
\end{tabular}
\vspace{8pt}
\centering
\caption{Ablation results on SyncRDE-60 $\times4$ (depth) with different selections of input modalities and modules. }
\label{tab:abl_depth}
\footnotesize
\setlength{\tabcolsep}{3pt}
\renewcommand{\arraystretch}{1}
\begin{tabular}{l c c c c c c c c}
\toprule
ID & R & D & E & ZR & Fus & RMSE $\downarrow$& AbsRel$\downarrow$& $\delta_{1.25}$ $\uparrow$\\
\midrule
V1  & \cmark& \cmark& \xmark& \cmark& F4& 0.0895& 0.2481& 0.8318\\
V2& \xmark& \cmark & \cmark & \cmark& F1& 0.0980& 0.2870& 0.8553\\
V4& \cmark & \cmark & \cmark & \xmark& F4& 0.0959& 0.2516& 0.8473\\
TL (RGB-D+E)& \cmark& \cmark& \cmark& -& -& 0.0601& 0.1599& 0.9014\\
\midrule
V5& \cmark& \cmark& \cmark& \cmark& F1& 0.0986& 0.2152& 0.8687\\
V6& \cmark& \cmark& \cmark& \cmark& F2
& 0.0897& 0.1701& 0.7712\\
 V7& \cmark& \cmark& \cmark& \cmark& F3& 0.0826& 0.1293&0.8681\\
\rowcolor{blue!10} 
UniRED& \cmark& \cmark& \cmark& \cmark& F4& 0.0570& 0.1031& 0.9346\\
\bottomrule
\end{tabular}
\end{table}

The quantitative results in the lower parts of the Tables \ref{tab:abl_rgb} and \ref{tab:abl_depth} reveal a clear and consistent trend.
F1 yields inferior performance because the two branches receive no explicit RGB-depth interaction, limiting cross-modal information exchange. 
F2 also performs poorly, suggesting that direct concatenation is insufficient for tri-modal fusion. 
Since RGB, depth, and event features exhibit different signal characteristics and encode distinct aspects of the scene, naive aggregation tends to entangle heterogeneous cues in a shared space, introducing cross-modal interference and weakening the discriminative structures needed for accurate interpolation.
F3 improves over F1 and F2 to some extent, but remains inferior to F4 because event features are not involved in the explicit RGB-depth fusion stage. Consequently, event cues can no longer directly guide cross-modal interaction at this stage, reducing the model’s ability to resolve fine-grained temporal misalignment near motion boundaries.
By contrast, the proposed fusion strategy F4 achieves the best performance for both RGB and depth interpolation, demonstrating that effective tri-modal fusion requires not only feature aggregation, but also event-guided interaction to establish accurate temporal and geometric correspondence between RGB and depth features.

\subsubsection{Ablation studies on refinement}
We respectively remove the Motion Basis Refinement (MB) and Z-axial Refinement (ZR) to measure their impact on the evaluation metrics. As shown in Table~\ref{tab:abl_rgb}, both depth-derived features and Motion Basis refinement notably improve the performance of RGB interpolation.
The former provides additional geometric context that complements appearance cues, while the latter introduces a structured motion prior that improves flow estimation in challenging regions, such as weak-texture areas and motion boundaries. Table~\ref{tab:abl_depth} further indicates that Z-axial Refinement brings a clear gain in depth estimation by explicitly correcting residual errors along the viewing direction, which are not well captured by in-plane flow alone.

\section{Conclusion}

In this paper, we introduce the new task of event-guided RGB-D video interpolation and present \systemname{}, the first unified framework for jointly interpolating RGB appearance and depth geometry from RGB, depth, and event inputs.
By integrating appearance, geometry, and temporally dense event cues within a tri-modal architecture, \systemname{} promotes effective cross-modal interaction and motion reasoning, yielding more faithful RGB reconstruction and more geometry-consistent depth recovery in highly dynamic scenes.
To facilitate this task, we established a standardized evaluation platform on two real-world benchmarks: a geometrically aligned RGB-D-Event version of VECtor and our newly collected synchronized RGB-D-Event dataset, SyncRDE-60.
Extensive experiments demonstrate that \systemname{} consistently outperforms existing frame-based, event-guided, and point-cloud-based baselines in both photometric quality and geometric accuracy, highlighting the effectiveness of jointly modeling appearance, geometry, and event dynamics for RGB-D video interpolation.

\bibliographystyle{IEEEtran}
\bibliography{IEEEabrv,reference}



 




\vfill

\end{document}